\pdfoutput=1

\documentclass[11pt]{article}

\usepackage[final]{acl}

\usepackage{times}
\usepackage{latexsym}
\usepackage{booktabs}
\usepackage{multirow}
\usepackage[T1]{fontenc}
\usepackage{rotating} 
\usepackage{lscape}    
\usepackage{graphicx}  

\usepackage[utf8]{inputenc}
\usepackage{enumerate}

\usepackage{microtype}

\usepackage{inconsolata}

\usepackage{graphicx}

\title{Climate-Eval: A Comprehensive Benchmark for NLP Tasks\\[1mm]
Related to Climate Change}

\author{
\large
\textbf{Murathan Kurfal{\i}}$^\dagger$$^{\P}$ \quad
\textbf{Shorouq Zahra}$^\dagger$$^{\S}$$^{\P}$ \quad \\
\textbf{Joakim Nivre}$^{\S}$$^{\P}$ \quad
\textbf{Gabriele Messori}$^{\S}$$^{\P}$\\[1mm]
$^\dagger$RISE Research Institutes of Sweden \\
$^{\S}$Uppsala University \\
$^{\P}$Swedish Centre for Impacts of Climate Extremes (climes)
}

\begin{document}
\maketitle
\begin{abstract}
Climate\-Eval is a comprehensive benchmark designed to evaluate natural language processing models across a broad range of tasks related to climate change. Climate\-Eval aggregates existing datasets along with a newly developed news classification dataset, created specifically for this release. This results in a benchmark of 25 tasks based on 13 datasets, covering key aspects of climate discourse, including text classification, question answering, and information extraction. Our benchmark provides a standardized evaluation suite for systematically assessing the performance of large language models (LLMs) on these tasks. Additionally, we conduct an extensive evaluation of open-source LLMs (ranging from 2B to 70B parameters) in both zero-shot and few-shot settings, analyzing their strengths and limitations in the domain of climate change. 

\end{abstract}

\section{Introduction}
Climate change represents one of the most pressing global challenges of our time, impacting every level of society, from international policy-making to everyday decisions. The importance of the topic is reflected in the vast amount of textual data generated, including a rich scientific literature as well as thousands of reports from corporations, government agencies and other organisations. Other data sources
include countless social media posts and news articles capturing all aspects of the debate on climate change, from urgent calls for action to widespread misinformation. However, this enormous supply of unstructured information is not in a format amenable to analysis, and manual processing is unfeasible due to the sheer volume of text at hand.

A possible solution to this challenge comes from the field of Natural language processing (NLP). NLP has shown remarkable progress in recent years, notably with LLMs achieving near human levels of performance across a variety of tasks. Thanks to their capacity to process textual data at scale, LLMs can help researchers process
large volumes of data for a wide range of applications, such as analyzing climate-related documents or social media posts \cite{el2021novel,upadhyaya2023multi}, structuring information about climate extremes from online texts into organized databases \cite{li2024using,madruga2025comprehensive}, and automatically detecting texts promoting climate change misinformation  \cite{zhang2024granular,cook2024generative}.  
LLMs are thus playing a key role in enabling the compilation and analysis of climate change information from textual sources. In this context, it is essential to assess their performance. Such assessment needs to be comprehensive and consider a wide variety of tasks, since the performance of LLMs is known to be highly domain-dependent \cite{ling2023domain}. In this paper, we aim to address this need by combining a newly-developed news classification dataset with existing datasets to create a unified benchmark that systematically evaluates the capabilities of LLMs across a vast array of climate-related NLP tasks. Our unified benchmark, Climate\-Eval, consists of 25 different tasks based on 13 datasets. It builds upon previous NLP benchmarking datasets, most notably Clima\-Bench \cite{spokoyny2023towards} and adds the following contributions:
\begin{itemize}
    \item We introduce a new topic classification dataset using news articles on climate-related topics from the Guardian newspaper.
    \item We compile diverse climate-related NLP tasks into a unified benchmark using LM Evaluation Harness \cite{eval-harness}\footnote{The benchmark can be accessed here: \url{https://github.com/NLP-RISE/ClimateEval}}, facilitating systematic evaluation of models on various aspects of climate change discourse.
    \item  We provide a comprehensive evaluation of a wide range of open-source LLMs on climate-related NLP tasks, highlighting these LLMs' strengths and limitations as well as looking into the different challenges they pose. 
    \item We provide a one-line evaluation setup to ensure accessibility and reproducibility for a wide range of users.
\end{itemize}

Through Climate\-Eval, we thus aim to facilitate NLP research by providing an easy-to-use setup enabling a comprehensive assessment of LLMs in climate change-related applications.
By covering diverse, manually annotated datasets, our benchmark offers insights into the applicability of LLMs to critical aspects of climate discourse, from stance detection to claim verification.


\section{Related Work}
Clima\-Bench \cite{spokoyny2023towards} offers a collection of datasets designed to evaluate NLP models on climate-related tasks, such as text classification and stance detection. To the best of our knowledge, it is the first effort to aggregate multiple datasets into a benchmark for NLP models in the climate change domain, laying the foundation for Climate\-Eval. In addition to curating existing datasets, Clima\-Bench introduces a new dataset, CDP (Carbon Disclosure Project), which is based on climate-related questionnaires filled out by different stakeholders. Clima\-Bench was used for the evaluation of Climate\-GPT \cite{thulke2024climategpt}, an LLM specifically fine-tuned for climate-related applications. However, the Climate\-GPT evaluation omitted or modified some of the Clima\-Bench tasks, including simplifying multi-class classification tasks into binary ones. Nonetheless, Climate\-GPT incorporates two additional datasets not included in Clima\-Bench in its evaluation suite, PIRA \cite{pirozelli2024benchmarks} and Exeter \cite{coan2021computer}, which we also include in our benchmark. \citet{fore2024unlearning} rely on question answering (QA) datasets to detect climate change misinformation in LLMs, specifically LLMs that have been intentionally injected with false climate information and subsequently made to unlearn it. 

The above work provides a context for evaluating LLMs on climate-related tasks. A key knowledge gap is that only a limited number of models have been benchmarked, and some tasks have suffered reductions (e.g., transforming multi-class classification into a binary classification problem) without considering the effect on performance when
transforming the task and target output
 in this fashion. Climate\-Eval complements and extends previous efforts by addressing this gap. Furthermore, the unified benchmark provides in unprecedented in breadth of both tasks and annotated data.
 
\begin{table*}[h]
\setlength{\tabcolsep}{2pt} 
    \centering
    \small
    \begin{tabular}{ l l r r r r }
        \toprule
        \textbf{Dataset (Source)} & \textbf{Task(s)} & \textbf{\# labels}& \textbf{Train} & \textbf{Dev} & \textbf{Test} \\
        \midrule
        ClimaText \cite{varini2020climatext} & Sentence classification  &2 & 121847 & 3918 & 5426  \\ 
        \hline & \\[-2ex]
        Climate-Stance \cite{vaid2022towards} & Stance classification &3 &2871 &354&355  \\ 
        \hline & \\[-2ex]
        Climate-Eng \cite{vaid2022towards} & Topic classification&  5 & 2871&354&355 \\ 
        \hline & \\[-2ex]
         Climate-FEVER \cite{diggelmann2020climate} & Claim verification  &3 & -&-&7675\\ 
        \hline & \\[-2ex]
        \multirow{3}{*}{SciDCC \cite{mishra2021neuralnere}} 
        & Topic classification by Title &20 &9231 &1154&1154 \\
        & Topic classification by Title \& Summary &20 &9231 &1154&1154 \\
        & Topic classification by Title \& Body &20 &9231 &1154&1154 \\ 
        \hline & \\[-2ex]
        \multirow{5}{*}{CLIMA-CDP \cite{spokoyny2023towards}} 
        & QA-Cities (answer relevance) 
        &2 &288418 &51018&55872 \\
        & QA-Corp. (answer relevance) 
        &2 &207450 &22044&29892 \\
        & QA-States (answer relevance)
        &2 &52287 &5814&6888 \\
        & Topic-Cities (topic classification) 
        &12 &46803 &8771&8984 \\
        \hline & \\[-2ex]
        \multirow{2}{*}{PIRA 2.0 MCQ \cite{pirozelli2024benchmarks}} 
        & PIRA with Context  &5 &1798 &225&227 \\
        & PIRA without Context &5 &1798 &225&227 \\ 
        \hline & \\[-2ex]
        \multirow{2}{*}{Exeter Misinformation \cite{coan2021computer}} 
        & Claim Detection  &6 &23436 &2605& 2904\\
        & Sub-claim Detection   &18 &23436 &2605& 2904 \\
        \hline & \\[-2ex]
        Climate-Change NER \cite{bhattacharjee2024indus} & Entity recognition &13&31633 &6366 &5775 \\ \hline & \\[-2ex]
        \multirow{5}{*}{CheapTalk \cite{bingler2023cheaptalk}} & Climate Detection  &2& 1300 & - & 400 \\
       
          & Climate Sentiment &3& 1000 & - & 320 \\
       
          & Climate Commitment & 2& 1300 & - & 400 \\
       
          & Climate Specificity &2& 1000 & - & 320 \\
      
          & TCFD Recommendations  &5& 1300 & - & 400 \\ 
        \hline & \\[-2ex]
        Net-Zero Reduction \cite{schimanski2023climatebert} &Paragraph Classification  &3 &3441 &-&-  \\ 
        \hline & \\[-2ex]
        Environmental Claims \cite{stammbach2022environmentalclaims} & Sentence Classification &2& 2400 & 300 & 300 \\ 
        \hline & \\[-2ex]

         \multirow{2}{*}{Guardian Climate News Corpus}  & Topic classification by Title &10&32138 &4017 &4018 \\
        & Topic classification by Body &10&  32138 &4017 &4018\\
        \bottomrule
    \end{tabular}
    \caption{Overview of the Climate\-Eval benchmark tasks, subtasks, and dataset sizes. The \textbf{\# labels} column gives the number of labels per task; for PIRA, it represents the number of answer choices per question, while for Climate-Change-NER, it corresponds to the total number of distinct entities annotated in the dataset.}
        \label{tab:climateEval}
\end{table*}

\section{The Climate\-Eval Benchmark}

In this section, we detail the datasets and the corresponding tasks included in the Climate\-Eval benchmark, along with the evaluation metrics we employ.

\subsection{Datasets and Tasks}
In this subsection, we first describe our new dataset, the Guardian Climate News Corpus, followed by the 
other datasets included in Climate\-Eval.

\begin{figure}[t]
    \centering
    \includegraphics[width=0.495\textwidth]{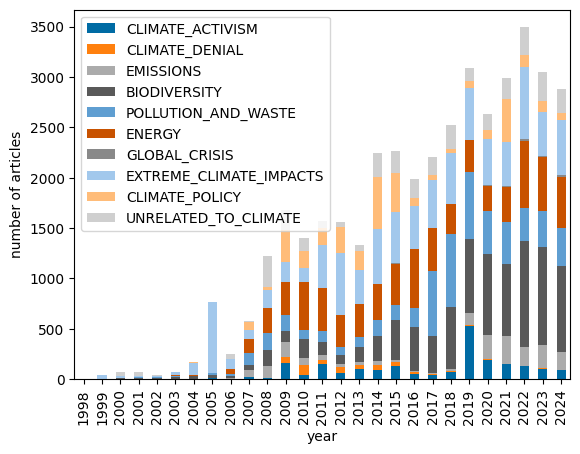}
    \caption{The distribution of articles by year in the Guardian Climate News Corpus, broken down by category.}
    \label{fig:guardian-year}
\end{figure}

\begin{itemize}


    
    \item \textbf{Guardian Climate News Corpus} 
    : A dataset containing climate-related and non-climate-related articles. These are assigned to nine climate-related categories (\textit{climate activism}, \textit{climate denial}, \textit{emissions}, \textit{biodiversity}, \textit{pollution and waste}, \textit{global crisis}, \textit{extreme climate impacts}, and \textit{climate policy}) and one \textit{unrelated to climate} category with articles sampled from diverse domains (e.g., sports, technology, gardening).\footnote{The benchmark and full details on the categories and tags can be foundon HuggingFace: \url{https://huggingface.co/datasets/NLP-RISE/guardian_climate_news_corpus}.}
    
    Data for each category is scraped from the Guardian's website\footnote{Using the Guardian Open Platform API \url{https://open-platform.theguardian.com/documentation/}}. The Guardian was chosen because it explicitly permits the use of its content for research and non-commercial purposes. Accordingly, we are able to freely generate and distribute this dataset. Articles are scraped based on having been assigned specific tags. These tags are used by the Guardian to taxonomize their own publications; each tag is usually comprised of a section and unique topic identifier, separated by a forward slash (e.g., the ``environment/flooding'' tag is assigned to articles covering flooding-related incidents in the Guardian's Environment section\footnote{\url{https://www.theguardian.com/uk/environment}}). This taxonomy has enabled us to curate a manual selection of tags that are relevant to each of the ten categories with ease.

     As example, an article that we categorize as falling under \textit{climate activism} is scraped based on a list of article tags that relate to climate activism (such as ``environment/school-climate-strikes'').  However, this article could also have been assigned other tags that describe it, such as ``australia-news/australian-education'' and ``world/extreme-weather'' -- the latter is a tag that falls under our "Extreme Climate Impacts" category. Since the second tag fall under another category we have defined, we remove this article from the pool so that each article can only belong to one category; otherwise, no action is taken. 
     After these are filtered out, any articles with a body shorter than 49 words or longer than 1,000 words are also dropped from the dataset, resulting in 40,173 articles in ten mutually-exclusive categories published any time between 1998 and 2024. Figure \ref{fig:guardian-year} shows the distribution of articles across years. 

     We derive two tasks from this dataset:
     \begin{enumerate}[i.]
        \item \textit{Topic classification by Title}, where multi-class classification is performed only on the title of the article; and
        \item \textit{Topic classification by Body}, where the same task is performed on the body of the article.
     \end{enumerate}

     We believe the Guardian Climate News Corpus is a valuable addition to existing resources, as it provides a large-scale, real-world news dataset focused on climate topics with fine-grained labels that capture diverse aspects within climate discourse. In contrast, existing datasets, such as SciDCC \cite{mishra2021neuralnere}, are not exclusively focused on climate change, but instead cover broader scientific categories, of which some are climate-related (e.g., ``Pollution'' and ``Hurricanes'').

     For simplicity, we have chosen to create a dataset to evaluate multi-class classification as each article belongs only to a single class. However, it is possible to reproduce this dataset with a different set of tags, with the option to not filter out articles sharing tags across categories, in turn creating a dataset more suitable for mutli-label classification.\footnote{The code used for scraping the dataset from the Guardian's Open Platform is available on GitHub: \url{https://github.com/NLP-RISE/extractguardian}}
    
    \item \textbf{ClimaText} \cite{varini2020climatext}: A  dataset with sentences from the web, Wikipedia, and public companies’ 10-K reports. Each sentence is labeled for whether it is related to climate change or not. Therefore, this datasets is suitable for the task of \textit{binary classification of sentences}, requiring models to distinguish texts relevant to climate change.
    
    \item \textbf{Climate-Stance} \cite{vaid2022towards}: A dataset consisting of 3,777 tweets posted during the 2019 United Nations Framework Convention on Climate Change COP (Conference of the Parties). The dataset is annotated for stance detection (classification), where each tweet is categorized into one of three classes: (i) being in \textit{favor} of climate change mitigation, (ii) being \textit{against} such measures, or (iii) taking an \textit{ambiguous} stance. 
        
    \item \textbf{Climate-Eng} \cite{vaid2022towards}: A dataset designed for multi-class topic classification, constructed from the same set of 3,777 tweets as the Climate-Stance dataset, but labeled according to one of five distinct topics: \textit{disaster}, \textit{ocean/water}, \textit{agriculture/forestry}, \textit{politics}, or \textit{general}.  
    
    \item \textbf{Climate-FEVER} \cite{diggelmann2020climate}: A claim verification dataset that consists of real-world claims about climate change. Each of the 1,535 claims is paired with five evidence sentences extracted from Wikipedia, which either \textit{support}, \textit{refute}, or \textit{provide insufficient information} about the claim. Climate-FEVER also includes a general label, determined by aggregating the ratings of individual claim-evidence pairs. However, in our benchmark, we model the task for this dataset as a three-way entailment problem where each claim is evaluated individually against each of its five evidence sentences, resulting in 7,675 claim-evidence pairs.

    \item \textbf{SciDCC} \cite{mishra2021neuralnere}: A multi-class classification dataset comprising news articles sourced from the Science Daily website, annotated with one of the possible 20 scientific categories (e.g., \textit{biology}, \textit{weather}, \textit{ozone holes}, \textit{endangered animals}). Each article includes a title, a summary, and a body section. In order to fully utilize the available information at different granularity, we model the task in three ways: 
    \begin{enumerate}[i.]
        \item \textit{Topic Classification by Title}, where classification is performed solely on the article title; 
        \item \textit{Topic Classification by Title \& Summary}, which utilizes both the title and the summary sections; and 
        \item \textit{Topic Classification by Title \& Body}, where the full article body is used, in addition to the title. 
    \end{enumerate}
     We aim for these complementary tasks to enable a comparative assessment of how much information is needed for an accurate classification performance.

    \item \textbf{CLIMA\--CDP} \cite{spokoyny2023towards}: A dataset derived from disclosure questionnaires collected and made available by Carbon Disclosure Project (CDP), an international non-profit organization.
    These questionnaires are completed by various stakeholders, including cities, corporations, and states. The dataset consists of responses to hundreds of unique questions related to climate impact, mitigation efforts and governance, and supports two distinct classification tasks: 

    \begin{enumerate}[i.]
        \item \textit{CDP\--QA} is a binary classification task that predicts whether a given report response correctly answers the questions posed. This task has three variants based on the type of stakeholder providing the response:\textit{ CDP\--QA\--Cities}, \textit{CDP\--QA\--Corporations}, and \textit{CDP\--QA\--States}, for  municipal, corporate and state-level responses, respectively.
        
        \item \textit{CDP\--Topic\--Cities} is a multi-class classification task where responses from city stakeholders are categorized into one of twelve predefined topics (e.g., \textit{climate hazards}, \textit{emissions}, \textit{energy}, \textit{food}). The task is limited to city responses since no annotations exist for other stakeholders.
    \end{enumerate}
      
  \item \textbf{PIRA 2.0 MCQ} \cite{pirozelli2024benchmarks}: A multiple-choice QA dataset constructed from a collection of scientific abstracts and United Nations reports, with a focus on climate-related topics such as oceanography, coastal ecosystems, and climate change impacts. Each instance consists of a question and five answer choices. Additionally, each question is accompanied by a supporting context that provides relevant information to guide the answer selection process. We divided the dataset into two sub-tasks:
    \begin{enumerate}[i.]
        \item \textit{PIRA with Context}, where models are expected to answer a question with the help of the accompanying context, like in an open-book exam setting; 
        \item \textit{PIRA without Context}, where models are required to answer the question without any supporting information. 
    \end{enumerate}

        By structuring the dataset in this manner, we aim to evaluate both retrieval-augmented and self-knowledge-based approaches to climate-related QA.
  
    \item \textbf{Exeter Misinformation} \cite{coan2021computer}: A dataset designed to detect climate misinformation by annotating text from prominent climate contrarian blogs and think tanks spanning over 20 years (1998–2020). The dataset is structured according to a two-level taxo\-nomy of climate contrarian claims. The first level consists of broad claim categories, such as: (1) \textit{Global warming is not happening}, which is further divided into more specific sub-claims, including (1.1) \textit{Ice isn’t melting}, or (1.2) \textit{Oceans are cooling}. To test different levels of misinformation detection, we define two sub-tasks: 
    \begin{enumerate}[i.]
        \item \textit{Claim Detection}, a classification task where texts are categorized into one of the six first-level claim labels; and
        \item \textit{Sub-Claim Detection}, a more fine-grained classification task where each text is assigned to its corresponding sub-claim category within the taxonomy. 
    \end{enumerate}
    These tasks allow for a detailed evaluation of how LLMs classify misinformation narratives in climate discourse at different granularities.
 
    \item \textbf{Climate\--Change NER} \cite{bhattacharjee2024indus}: A \textit{named entity recognition} (NER) dataset constructed from 534 scientific abstracts sourced from the Semantic Scholar Academic Graph \cite{kinney2023semantic}. The dataset was collected using a set of climate-related keywords (e.g., \textit{wildfire}, \textit{floods}) to ensure relevance to climate science. Each abstract is annotated for entity types that are specific to the climate discourse, such as \textit{greenhouse gases}, \textit{climate hazards}, \textit{climate-impacts}, \textit{climate mitigations}. The associated task is a token classification task, where the goal is to identify tokens corresponding to these entities. NER is a crucial preprocessing step for various downstream applications that aim to extract structured information from unstructured text \cite{li2024using}. Hence, a model's performance on climate-specific NER is a strong indicator of its usefulness for information extraction in climate discourse \cite{li2020survey}.

\begin{table*}[h]\centering\scriptsize

\begin{tabular}{l cc | cc | cc | cc | cc | cc | cc | cc | cc}
\toprule
\textbf{Task} & \multicolumn{2}{c}{\textbf{\shortstack{Gemma-2\\ 2B}}} & \multicolumn{2}{c}{\textbf{\shortstack{Llama-2\\ 7B}}} & \multicolumn{2}{c}{\textbf{\shortstack{Llama-2\\ 13B}}} & \multicolumn{2}{c}{\textbf{\shortstack{Climate\\GPT-7B}}} & \multicolumn{2}{c}{\textbf{\shortstack{Climate\\GPT-13B}}} & \multicolumn{2}{c}{\textbf{\shortstack{Qwen-2.5\\ 7B}}} & \multicolumn{2}{c}{\textbf{\shortstack{Llama-3.1\\ 8B}}} & \multicolumn{2}{c}{\textbf{\shortstack{Mistral\\ 24B}}} & \multicolumn{2}{c}{\textbf{\shortstack{Llama-3.3\\ 70B}}} \\ \midrule
            & \multicolumn{1}{c}{0} & \multicolumn{1}{c}{5} & \multicolumn{1}{c}{0} & \multicolumn{1}{c}{5} & \multicolumn{1}{c}{0} & \multicolumn{1}{c}{5} & \multicolumn{1}{c}{0} & \multicolumn{1}{c}{5} & \multicolumn{1}{c}{0} & \multicolumn{1}{c}{5} & \multicolumn{1}{c}{0} & \multicolumn{1}{c}{5} & \multicolumn{1}{c}{0} & \multicolumn{1}{c}{5} & \multicolumn{1}{c}{0} & \multicolumn{1}{c}{5} & \multicolumn{1}{c}{0} & \multicolumn{1}{c}{5} \\ \cmidrule(lr){2-19}
   CDP-QA-Cities & .46 & .60 & .62 & \textbf{.69} & .37 & .55 & .48 & .57 & .59 & .57 & .61 & .62 & .57 & .66 & \underline{.68} & \underline{.68} & .61 & \textbf{.69} \\ 
CDP-QA-Corp. & .46 & .59 & .62 & \underline{.67} & .32 & .55 & .40 & .57 & .58 & .58 & .60 & .61 & .51 & .64 & \underline{.67} & \textbf{.68} & .60 & \textbf{.68} \\ 
CDP-QA-States & .46 & .61 & .68 & \underline{.71} & .37 & .56 & .48 & .59 & .60 & .58 & .62 & .63 & .54 & .66 & \textbf{.72} & .68 & .59 & .70 \\ 
CDP-Topic-Cities & .28 & .34 & .32 & \underline{.35} & \underline{.35} & \underline{.35} & .33 & \underline{.35} & .30 & \underline{.35} & .30 & \underline{.35} & .31 & \textbf{.37} & .34 & .32 & \underline{.35} & \underline{.35} \\ 
Climate Commit. & .42 & .63 & .66 & \textbf{.71} & .56 & .63 & .63 & .65 & .61 & .60 & .56 & \underline{.69} & .68 & \textbf{.71} & \textbf{.71} & \textbf{.71} & .68 & .68 \\ 
Climate Detection & .20 & .70 & .65 & .69 & .61 & .60 & .59 & .66 & .62 & .68 & .53 & .73 & .50 & \underline{.76} & .69 & \textbf{.78} & .71 & \underline{.76} \\ 
Climate Eng & .37 & .46 & .52 & \underline{.59} & .37 & .50 & .52 & .53 & .53 & .50 & .52 & .54 & .55 & \underline{.59} & \textbf{.60} & \underline{.59} & .57 & \underline{.59} \\ 
Climate NER & .11 & .19 & .14 & .20 & .10 & .17 & .09 & .17 & .07 & .14 & .04 & .18 & .16 & .21 & .23 & \textbf{.30} & .20 & \underline{.27} \\ 
Climate Sentiment & .49 & .64 & .72 & \underline{.74} & .44 & .71 & .60 & .70 & .61 & .65 & .57 & .63 & .54 & .72 & \textbf{.75} & .72 & .73 & \textbf{.75} \\ 
Climate Specificity & .39 & .66 & .72 & .69 & .54 & .60 & .51 & .64 & .48 & .60 & .53 & .68 & .60 & .73 & \underline{.75} & \textbf{.79} & \underline{.75} & \textbf{.79} \\ 
Climate Stance & .13 & .50 & .30 & .54 & .20 & .47 & .16 & .48 & .11 & .37 & .14 & .42 & .06 & .56 & .18 & \underline{.61} & .26 & \textbf{.64} \\ 
Climate-Fever & .33 & .34 & \textbf{.57} & .48 & .40 & .34 & .35 & .41 & .25 & .45 & .41 & .55 & .52 & .51 & \underline{.56} & .50 & .51 & .55 \\ 
Climatext Sent Clf. & .39 & .64 & .56 & .62 & .58 & .67 & .63 & \underline{.70} & .58 & .61 & .54 & .68 & .47 & .68 & .57 & .68 & .63 & \textbf{.71} \\ 
Env. Claims & .53 & .77 & .80 & .81 & .61 & .79 & .66 & .78 & .80 & .75 & .64 & .81 & .75 & \underline{.83} & .75 & .82 & \textbf{.85} & \underline{.83} \\ 
Exeter Claim & .14 & .41 & .43 & .48 & .17 & .30 & .20 & .38 & .26 & .37 & .34 & .41 & .35 & .46 & .55 & \textbf{.61} & .56 & \underline{.59} \\ 
Exeter Sub-Claim & .22 & .27 & .48 & .51 & .05 & .13 & .15 & .25 & .24 & .25 & .33 & .40 & .39 & .47 & \underline{.61} & \textbf{.63} & .59 & .59 \\ 
Guardian Body & .39 & .31 & .48 & \underline{.54} & .29 & .19 & .31 & .21 & .30 & .25 & .33 & .45 & .47 & .49 & \textbf{.58} & .00 & .00 & .00 \\ 
Guardian Title & .35 & .35 & .45 & \underline{.54} & .40 & .41 & .38 & .00 & .46 & .48 & .45 & .00 & .42 & .50 & .51 & .53 & \textbf{.57} & .00 \\ 
Net-Zero Reduction & .22 & .44 & .38 & .45 & .46 & .84 & .39 & \underline{.86} & .23 & .83 & .43 & .85 & .48 & .43 & .28 & .47 & .60 & \textbf{.89} \\ 
Pira W/ Ctx. & .86 & .86 & \underline{.95} & .94 & .74 & .70 & .87 & .88 & .86 & .87 & .93 & .94 & .93 & \underline{.95} & .94 & \textbf{.96} & .94 & .92 \\ 
Pira W/O Ctx. & .64 & .62 & .84 & .84 & .44 & .52 & .58 & .67 & .67 & .74 & .70 & .81 & .69 & .80 & .80 & \textbf{.89} & \textbf{.89} & \underline{.88} \\ 
SciDCC Title & .16 & .19 & .27 & \underline{.28} & .18 & .23 & .08 & .20 & .16 & .17 & .16 & .24 & .15 & .23 & \underline{.28} & \textbf{.33} & .25 & \textbf{.33} \\ 
SciDCC Title Body & .10 & .21 & .20 & .23 & .13 & .14 & .08 & .17 & .13 & .11 & .17 & .18 & .18 & .23 & .26 & \textbf{.32} & .26 & \underline{.31} \\ 
SciDCC Title Sum. & .10 & .19 & .25 & .27 & .13 & .25 & .12 & .25 & .17 & .16 & .20 & .25 & .20 & .27 & .28 & \underline{.33} & .27 & \textbf{.34} \\ 
Tcfd Recommend. & .15 & .31 & .43 & .46 & .24 & .30 & .24 & .30 & .16 & .29 & .29 & .37 & .26 & .45 & .50 & \underline{.52} & \textbf{.53} & .48 \\ 

            \bottomrule
        \end{tabular}
\caption{Evaluation results for various models across different few-shot experiments. The numbers indicate the models' performance for each task in F1-macro, except for PIRA (see Section \ref{sec:metrics}). Numbers in boldface are the highest performing, whereas underlined numbers are the second highest for that task. For shortened task identifiers, see Table \ref{tab:climateEvalShortened}.}
    \label{tab:results}
\end{table*}

\item{\textbf{CheapTalk} \citep{bingler2023cheaptalk}}: A dataset consisting of corporate disclosures, focusing on how companies communicate climate-related information. Unlike the previously described datasets, primarily dealing with either academic discourse or social media posts, this dataset uncovers
corporations' climate communication strategies. The following five tasks are performed on it:

\begin{enumerate}[i.]
    \item \textit{Climate Detection}: A binary classification task to determine whether a given text passage discusses climate-related topics. This dataset serves as a filtering mechanism to extract climate-relevant content from corporate reports.

    \item \textit{Climate Sentiment}: A sentiment analysis task that categorizes climate-related statements based on their tone. Each passage is labeled as highlighting risks, emphasizing opportunities, or maintaining a neutral stance on climate change.
    
    \item \textit{Climate Commitments}: A classification task that determines whether a corporate disclosure contains a climate-related commitment.
    Texts are labeled as \textit{commitment-yes} if they explicitly state planned or ongoing climate actions and \textit{commitment-no} if they do not reference concrete climate actions.
    
    \item \textit{Climate Specificity}: A binary classification task assessing the specificity of corporate climate commitments. A passage is labeled as \textit{specific} if containing ``detailed performance information, details of actions, or tangible and verifiable targets'' \cite{bingler2023cheaptalk}, or \textit{non-specific} if it is vague.
    
    \item \textit{TCFD Recommendations}: A multi-class classification task assessing corporate disclosures against the guidelines of the Task Force on Climate-related Financial Disclosures (TCFD), an international framework that standardizes the reporting of climate-related financial risks and opportunities.\footnote{\url{https://www.fsb-tcfd.org}} Each text is labelled by one of the four TCFD recommendation categories (\textit{governance}, \textit{strategy},  \textit{risk management}, and \textit{metrics and targets}) or as \textit{none} if none applies.
\end{enumerate}


\item \textbf{Net-Zero Reduction} \cite{schimanski2023climatebert}: A dataset comprising 3,517 expert-annotated paragraph samples designed to detect and assess net-zero and emission reduction targets in corporate, national, and regional communications. Each sample is labeled as \textit{net-zero target} (commitment to net-zero emissions), \textit{reduction target} (commitment to emission reduction without full net-zero), or \textit{no target} (no explicit reduction commitment). This dataset is used for multi-class classification.

\item \textbf{Environmental Claims} \cite{stammbach2022environmentalclaims}: A dataset comprising 3,000 sentences from corporate sustainability reports, earnings calls, and annual reports by publicly listed companies that are expert-annotated for environmental claims. An environmental claim any statement suggesting that a product, service, or company is environmentally friendly. Each sentence is labeled as containing an environmental claim or not. This dataset is used for a binary sentence classification task.
    
\end{itemize}

Most of the datasets in ClimateEval were developed in the pre-LLM era and were not designed with prompting-based evaluation in mind. Moreover, prior work has used some datasets in different ways — for example, ClimateGPT modeled Climate-FEVER as a binary classification task (not three-way). To address this lack of standardization, ClimateEval provides a unified evaluation benchmark that standardizes task formulations, label sets, and prompts. Each task is paired with a suitable prompt and, where applicable, modeled at multiple granularities (e.g., title-only vs. full-body classification in SciDCC). This unification ensures that different models are evaluated under consistent conditions, enabling reproducible comparisons across tasks.

\subsection{Evaluation Setup and Metrics} \label{sec:metrics}

ClimateEval is implemented using the LM Evaluation Harness library \cite{eval-harness}, which provides an easy-to-use infrastructure for evaluating language models on a wide range of tasks. Each task in the benchmark is defined through a YAML-formatted configuration file, specifying the input format, prompt template, expected output and the target metric. The benchmark can be executed with a single command, allowing for efficient and standardized evaluation across diverse models and tasks.

We evaluate model performance primarily using accuracy and macro-averaged F1-score for all classification tasks. Macro F1 is particularly important due to class imbalances in several datasets where there are as many as 20 labels with skewed distributions (see Appendix \ref{appendix:stats} for details). Macro F1 averages scores across all classes equally to ensure a balanced assessment by preventing frequent labels from biasing the results.

For Climate-Change-NER, the only sequence labeling task in our benchmark, we compute precision, recall, and F1-score based on the entity-type and entity-span pairs. Each entity-type is evaluated independently, and a model’s prediction is considered correct if it correctly identifies an entity within the set of gold entities for that type. If a gold-standard entity-type contains multiple entities and the model predicts only a subset, we count each correctly identified entity as a true positive, while missing entities contribute to false negatives. We report only exact matches, where an entity is correct only if both the span and type match perfectly.

For multiple-choice QA (MCQA) tasks (e.g., PIRA 2.0 MCQ), we report exact match accuracy, as F1-score is not meaningful in this context. In MCQA tasks, each question has one correct answer, and the model selects from arbitrary option labels (e.g., A, B, C), which are not semantic classes. Since the model selects a single arbitrary option and only one answer is correct, precision and recall are not meaningful, making exact match accuracy the appropriate metric.

\section{Evaluation of Open-Source LLMs}

We report the performance of a range of open source baseline LLMs with varying sizes, ranging from 2B to 70B, in both zero-shot and 5-shot scenarios. The mid-sized models (Mistral 24B and Llama3.3-70B) are loaded in 4-bit quantization whereas the other models were run in half-precision FP16. 

For classification tasks, the log-likelihoods of each possible label are calculated, and the label with the highest likelihood is selected as the model's prediction. For the generation task, Climate Change NER, the model is simply prompted to generate the corresponding JSON file.

The baseline LLMs that we use are: Gemma-2 (2B) \cite{team2024gemma}, Qwen-2.5 (7B) \cite{yang2024qwen2}, Climate\-GPT (both 7B and 13B) \cite{thulke2024climategpt}; Llama-2 (both 7B and 13B); Llama-3.1 (8B), Llama-3.3 (70B) \cite{dubey2024llama} and Mistral (24B). The baseline models' performances are reported in Table \ref{tab:results} in both zero-shot and five-shot settings. 

\subsection{Zero-shot vs. Few-shot Performance}
\begin{figure}[t]
    \centering
    \includegraphics[width=0.48\textwidth]{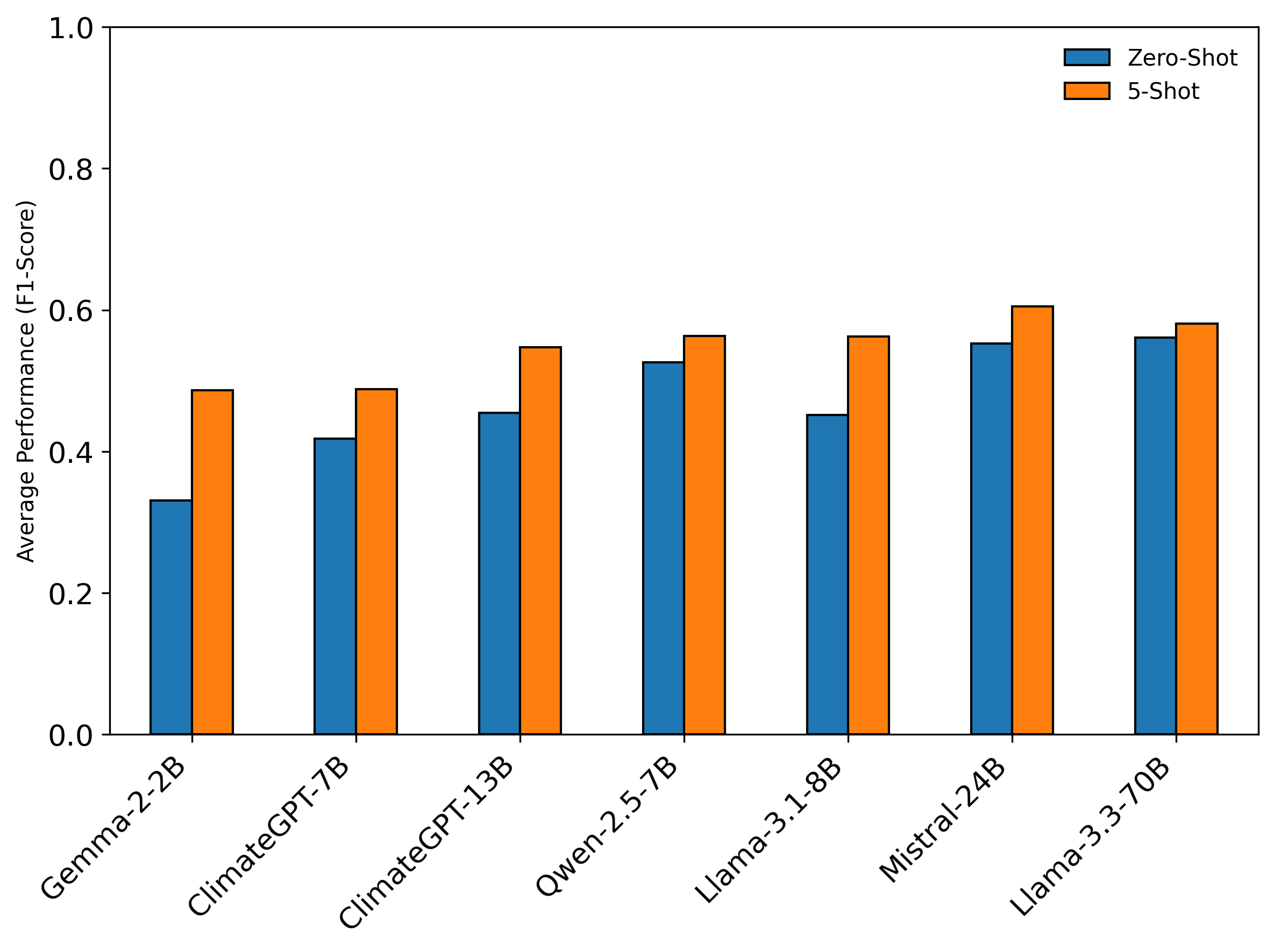}
    \caption{Average model performance across all tasks in Zero-Shot and 5-Shot settings.}
    \label{fig:avg}
\end{figure}

Across the benchmark, few-shot prompting consistently improves performance (Figure \ref{fig:avg}). The most significant gains are observed in Climate Stance ($+0.26$ for ClimateGPT-7B to $+0.50$ for Llama-3.1-8B) and Net-Zero Reduction ($+0.15$ for Qwen-2.5-7B to $+0.44$ for Gemma-2-2B). Both tasks rely on understanding specialized classification taxonomies, whether for categorizing social media discourse or analyzing policy documents. The improvement diminish as the number of labels increases. The Exeter Sub-claim Detection and SciDCC tasks exhibit minimal gains ($0.03$ on average), suggesting that when models must choose between a high number of categories, few-shot prompting does not provide sufficient guidance. 

Another set of tasks that do not benefit from in-context learning includes Climate-FEVER (fact verification) and PIRA (multiple-choice QA), which show no improvement. Unlike standard classification tasks, these tasks require external knowledge to be successfully carried out, limiting the efficiency of few-shot prompting. As each instance in these datasets depends on specific knowledge, the lack of improvement is unsurprising.

\begin{table}[t]
\centering
\small
\begin{tabular}{lccc}
\toprule
\textbf{Task} & \textbf{0-Shot} & \textbf{5-Shot} & \textbf{Diff.} \\
\midrule
CDP-QA-Cities & .55 (.10) & .63 (.06) & .07 \\
CDP-QA-Corp. & .53 (.11) & .62 (.05) & .09 \\
CDP-QA-States & .56 (.11) & .64 (.06) & .07 \\
CDP-Topic-Cities & .32 (.02) & .35 (.01) & .03 \\
Climate Commit. & .61 (.09) & .67 (.04) & .06 \\
Climate Detection & .57 (.15) & .71 (.06) & .14 \\
Climate Eng & .50 (.08) & .54 (.05) & .04 \\
Climate NER & .13 (.06) & .20 (.05) & .08 \\
Climate Sentiment & .60 (.11) & .70 (.04) & .09 \\
Climate Specificity & .58 (.13) & .69 (.07) & .10 \\
Climate Stance & .17 (.07) & .51 (.09) & .34 \\
Climate-Fever & .43 (.11) & .46 (.08) & .03 \\
ClimaText Sent Clf. & .55 (.08) & .67 (.03) & .12 \\
Env. Claims & .71 (.10) & .80 (.03) & .09 \\
Exeter Claim & .33 (.16) & .45 (.10) & .11 \\
Exeter Sub-Claim & .34 (.19) & .39 (.17) & .05 \\
Guardian Body & .42 (.13) & .41 (.18) & -.01 \\
Guardian Title & .44 (.07) & .48 (.08) & .04 \\
Net-Zero Reduction & .39 (.13) & .67 (.22) & .29 \\
Pira W/ Ctx. & .89 (.07) & .89 (.08) & .00 \\
Pira W/O Ctx. & .70 (.14) & .75 (.13) & .06 \\
SciDCC Title Sum. & .19 (.07) & .26 (.06) & .07 \\
SciDCC Title & .19 (.07) & .24 (.06) & .06 \\
SciDCC Title Body & .17 (.06) & .21 (.07) & .04 \\
TCFD Recommend. & .31 (.14) & .39 (.09) & .07 \\
\toprule
Average & .45 & .53  &  .09 \\
\bottomrule
\end{tabular}
\caption{Comparison of average model performance across tasks in 0-shot and 5-shot settings. Standard deviations are provided within parentheses. The last column shows the performance difference between the zero-shot and five-shot experiments. For shortened task identifiers, see Table \ref{tab:climateEvalShortened}.}
\label{tab:avg_performance}
\end{table}

\subsection{Analysis of tasks}
Among the 25 tasks in our benchmark, certain tasks consistently emerge as more challenging than others (Table \ref{tab:avg_performance}). SciDCC and Climate-Change-NER exhibit particularly poor performance, with F1-scores below 0.34 across models, even with few-shot prompting. The poor performance on SciDCC can be attributed to the large number of classes (20 in total), making the assignment more difficult. Surprisingly enough, classification of news texts based solely on their titles only leads to a minor performance loss compared to using the full article. This may indicate that titles provide sufficient cues regarding the content of articles.

Climate-Change NER, on the other hand, is the only non-classification task in the benchmark, requiring models to correctly identify climate-related named entities in scientific abstracts. This task involves both span detection and entity type classification, where entity types are domain-specific\footnote{For the full list of entities, see: \url{https://huggingface.co/datasets/ibm-research/Climate-Change-NER}} and differ significantly from standard NER labels such as "location" or "organization". Without specific training/fine-tuning on climate-specific entity annotations, it is understandable that general-purpose LLMs struggle to accurately extract these entities.

In contrast, the LLMs perform strongly on several tasks in the benchmark, even in the zero-shot setting. The performance on PIRA improves by $0.2$ points on average in the zero-shot setting when relevant context is provided, showcasing the LLMs' ability to identify and utilize relevant parts of additional knowledge for question answering (QA). However, the overall high performance on PIRA even without any context suggests that the task may be relatively easy, highlighting the need for a more challenging QA dataset.

Perhaps the most surprising finding is the relatively poor performance of the models on the CheapTalk dataset tasks. Intuitively, one might expect tasks like Climate Detection (whether or not a text is climate-related) or Climate Sentiment (classifying tone as risk, opportunity or neutral) to be straightforward for advanced LLMs. Yet, our results show that even the best models achieve F1 scores below 0.8 on these tasks, indicating that these models experience challenges in understanding the climate discourse.

\subsection{Impact of In-domain Training} 
\label{sec:domain} 

Climate\-GPT presents a unique opportunity to evaluate the impact of in-domain training for climate-related NLP, as it is an LLM developed through continuous pre-training of Llama-2 models on 4.2B climate-focused tokens and instruction-tuned on expert-curated datasets \cite{thulke2024climategpt}. To gain insights regarding the benefits of domain adaptation, we compare Climate\-GPT against its base Llama-2 models (Table \ref{tab:climate_vs_llama}). 

Table \ref{tab:avg_performance} shows the performance differences between these models. On average, Climate\-GPT exhibits a slight improvement over Llama-2 by $\leq 0.05$. The largest gains are seen in Exeter Sub-claim, CDP\--QA, and PIRA With Context. However, the improvements are inconsistent, with some tasks, including Climate\--Change NER, Climate Stance, and Net\--Zero Reduction, showing negligible or negative differences. The limited impact of in-domain training across tasks suggests that, while climate-focused continuous pre-training provides benefits in specific cases, the performance gains are not uniform across the tasks in Climate\-Eval.

\begin{table}[t]
\centering
\small

\begin{tabular}{lcc|cc}
\toprule
\textbf{Task} & \multicolumn{2}{c|}{\textbf{7B}} & \multicolumn{2}{c}{\textbf{13B}} \\
 & \textbf{0} & \textbf{5} & \textbf{0} & \textbf{5} \\
\midrule
CDP-QA-Corp. & \textbf{\phantom{-}0.26} & \textbf{\phantom{-}0.02} & \textbf{0.20} & \textbf{\phantom{-}0.04} \\
CDP-QA-States & \textbf{\phantom{-}0.23} & \textbf{\phantom{-}0.02} & \textbf{0.13} & \textbf{\phantom{-}0.04} \\
CDP-QA-Cities & \textbf{\phantom{-}0.22} & \textbf{\phantom{-}0.03} & \textbf{0.13} & \textbf{\phantom{-}0.05} \\
CDP-Topic-Cities & -0.05 & \phantom{-}0.00 & -0.03 & \phantom{-}0.00 \\
ClimaText Sent. Clf. & -0.00 & -0.06 & -0.10 & -0.02 \\
Climate NER & -0.03 & -0.04 & -0.04 & \phantom{-}0.00 \\
Climate Commit. & \textbf{\phantom{-}0.05} & -0.03 & -0.07 & \textbf{\phantom{-}0.04} \\
Climate Detection & \textbf{\phantom{-}0.01} & \textbf{\phantom{-}0.07} & -0.06 & \textbf{\phantom{-}0.08} \\
Climate Eng & \textbf{\phantom{-}0.16} & \phantom{-}0.00 & -0.00 & \phantom{-}0.00 \\
Climate-FEVER & -0.15 & \textbf{\phantom{-}0.11} & \textbf{\phantom{-}0.07} & \textbf{\phantom{-}0.14} \\
Climate Sentiment & \textbf{\phantom{-}0.17} & -0.06 & -0.03 & -0.06 \\
Climate Specificity & -0.06 & \textbf{\phantom{-}0.01} & \textbf{\phantom{-}0.02} & \textbf{\phantom{-}0.04} \\
Climate Stance & -0.09 & -0.10 & -0.01 & -0.06 \\
Env. Claims & \textbf{\phantom{-}0.19} & -0.05 & -0.02 & \textbf{\phantom{-}0.03} \\
Exeter Claim & \textbf{\phantom{-}0.09} & \textbf{\phantom{-}0.07} & \textbf{\phantom{-}0.14} & \textbf{\phantom{-}0.03} \\
Exeter Sub-claim & \textbf{\phantom{-}0.19} & \textbf{\phantom{-}0.12} & \textbf{\phantom{-}0.19} & \textbf{\phantom{-}0.16} \\
Guardian Body & \textbf{\phantom{-}0.02} & \textbf{\phantom{-}0.06} & \textbf{\phantom{-}0.02} & \textbf{\phantom{-}0.24} \\
Guardian Title & \textbf{\phantom{-}0.06} & \textbf{\phantom{-}0.07} & \textbf{\phantom{-}0.07} & \textbf{\phantom{-}0.08} \\
Net-Zero Reduction & -0.24 & -0.01 & \textbf{\phantom{-}0.04} & -0.01 \\
PIRA w/ Ctx. & \textbf{\phantom{-}0.12} & \textbf{\phantom{-}0.17} & \textbf{\phantom{-}0.06} & \textbf{\phantom{-}0.06} \\
PIRA w/o Ctx. & \textbf{\phantom{-}0.23} & \textbf{\phantom{-}0.23} & \textbf{\phantom{-}0.12} & \textbf{\phantom{-}0.15} \\
SciDCC Title & -0.02 & -0.07 & \textbf{\phantom{-}0.08} & \textbf{\phantom{-}0.04} \\
SciDCC Title Sum. & \textbf{\phantom{-}0.04} & -0.10 & \textbf{\phantom{-}0.08} & \phantom{-}0.00 \\
SciDCC Title Body & -0.00 & -0.02 & \textbf{\phantom{-}0.08} & \textbf{\phantom{-}0.01} \\
TCFD Recommend. & -0.08 & -0.02 & \textbf{\phantom{-}0.05} & \textbf{\phantom{-}0.07} \\

\midrule
\textbf{Average} & \textbf{\phantom{-}0.05} & \textbf{\phantom{-}0.01} & \textbf\phantom{-}{0.04} & \textbf{\phantom{-}0.04} \\
\bottomrule
\end{tabular}
\caption{Comparison of Climate\-GPT against its' base Llama models in \textbf{0}-shot and \textbf{5}-shot settings. The values represent the difference where positive values (highlighted in bold) indicate better performance of Climate\-GPT over Llama.}
\label{tab:climate_vs_llama}
\end{table}



\section{Conclusion}

We have presented Climate\-Eval, a comprehensive benchmark for climate change NLP, encompassing 13 datasets and 25 tasks that cover a wide range of climate-related language understanding tasks. By unifying these diverse tasks into a single framework, Climate\-Eval enables systematic assessment of how well current LLMs perform in the domain of climate change. Our evaluation of some widely-used open-source LLMs revealed systematic patterns: few-shot prompting generally improves performance, but certain text classification tasks such as claim detection or climate-specific NER remain challenging. We hope that Climate\-Eval will serve as a valuable resource for the NLP community and facilitate future research on evaluating and improving LLMs for climate change-related applications.

\section*{Limitations}
We acknowledge that ClimateEval is currently limited to English, which restricts its applicability to multilingual climate discourse. Due to computational constraints our evaluation focuses on mid-sized open-source models, ranging from 2B to 70B parameters, with larger models tested under 4-bit quantization. This may introduce some performance degradation and may not reflect their optimal performance. Additionally, commercial models such as GPT-4 and Claude are not included in our experiments due to budget constraints. Finally, the benchmark largely consists of \textit{n-way} classification tasks, with the only exception of Climate-Change NER. This was partly driven by available climate-relevant datasets, which are predominantly classification-oriented. Future work should focus on enabling assessment of generative tasks such as information extraction, text generation or multimodal classification.

\section*{Acknowledgments}
Swedish Research Council Vetenskapsr\aa det grant nos. 2022-03448, 2022-06599 and 2022-02909. European Commission's Horizon Europe research and innovation programme European Research Council grant no. 101112727.


\bibliography{custom}
\newpage

\appendix

\section{Models} \label{appendix:models}
Table \ref{tab:hf} lists the baseline models used in our paper, along with their corresponding repository names on \url{https://huggingface.co/}. 

\begin{table}[ht] 
\centering
\small
\begin{tabular}{l|l} \hline  \\[-1ex]
\textbf{Model Name} & \textbf{HuggingFace Repository} \\ \hline \\[-1.5ex]
Gemma-2-2B & google/gemma-2-2b-it \\[0.5ex]
Qwen-2.5-7B & Qwen/Qwen2.5-7B \\[0.5ex]
Llama-2-7B & meta-llama/Llama-2-7b-chat-hf \\[0.5ex]
Llama-2-13B & meta-llama/Llama-2-13b-chat-hf \\[0.5ex]
ClimateGPT-7B & climategpt/climategpt-7b \\[0.5ex]
ClimateGPT-13B & climategpt/climategpt-13b \\[0.5ex]
Llama-3.1-8B & meta-llama/Llama-3.1-8B-Instruct \\[0.5ex]
Mistral-24B & mistralai/Mistral-24B-Instruct \\[0.5ex]
Llama-3.3-70B & meta-llama/Llama-3.3-70B-Instruct \\[0.5ex] \hline 

\end{tabular} \caption{HuggingFace repository names of the baseline models used in our evaluation.} \label{tab:hf} \end{table}

\section{CO2 Emission Related to Experiments}

Experiments were conducted using a private infrastructure, which has a carbon efficiency of 0.432 kgCO$_2$eq/kWh. A cumulative of 240 hours of computation was performed on hardware of type A100 SXM4 80 GB (TDP of 400W). Total emissions are estimated to be 41.47 kgCO$_2$eq of which 0 percent were directly offset. Estimations were conducted using the \href{https://mlco2.github.io/impact#compute}{MachineLearning Impact calculator} presented in \citet{lacoste2019quantifying}.

\section{Label distribution} \label{appendix:stats}
\begin{figure*}[b]
    \centering
    \includegraphics[width=\textwidth]{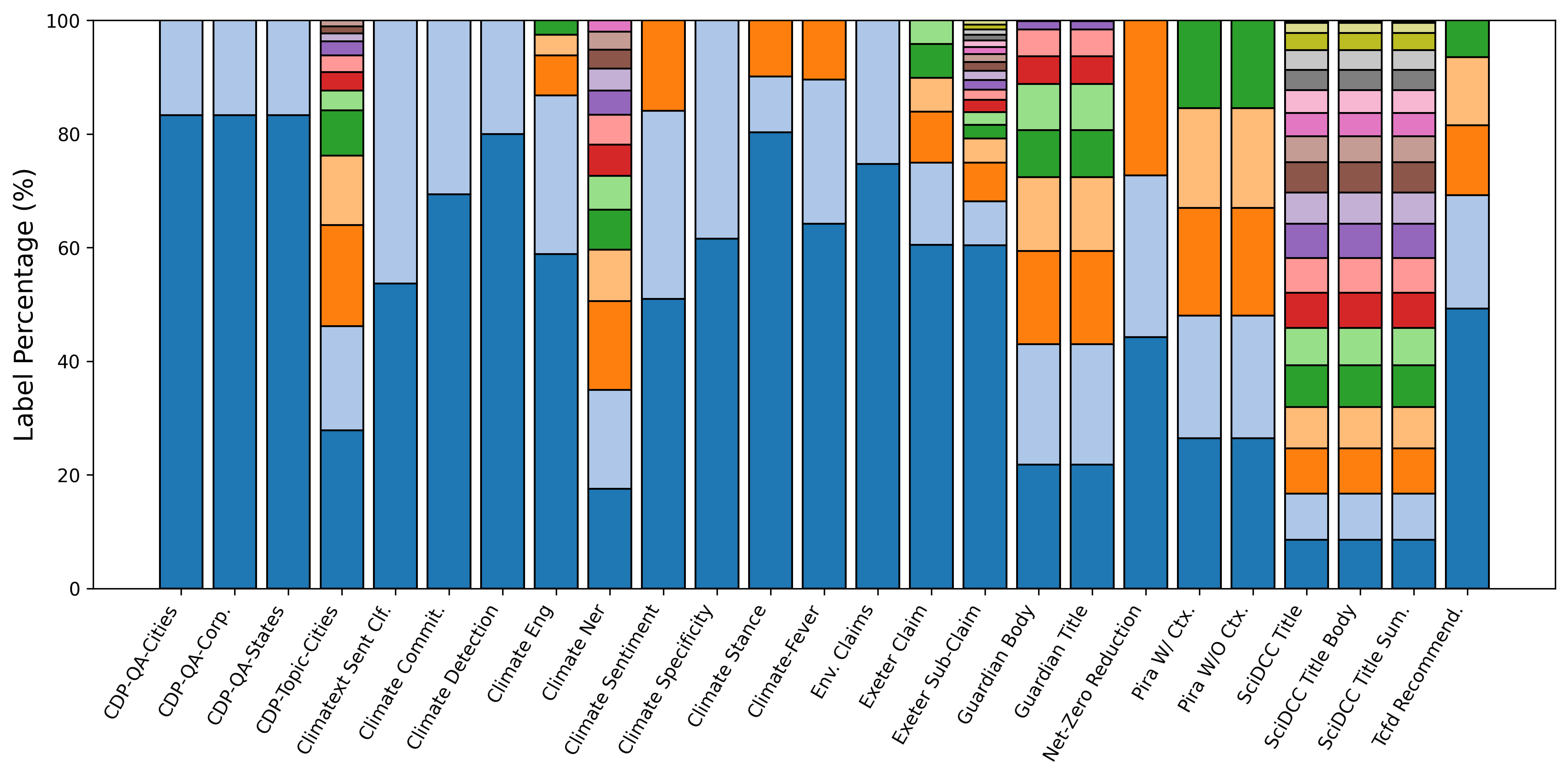}  
    \caption{The normalized distribution of labels in each task across all sets (train, development, and test).}    \label{fig:label_distributions}
\end{figure*}

To provide further insights into the datasets, we visualize the label distribution for each test set in Figure \ref{fig:label_distributions}. Given the wide size range of the test sets (between 300 and 55872), we present the normalized label distributions, where each stack in a bar represents the percentage of a label within the corresponding task's test set.

\begin{table*}[h]
\setlength{\tabcolsep}{2pt} 

    \centering
    \small
    \begin{tabular}{ l l l }
        \toprule
        \textbf{Dataset (Source)} & \textbf{Task(s)} & \textbf{Shortened identifier} \\
        \midrule
        ClimaText \citep{varini2020climatext} & Sentence classification  & ClimaText Sent. Clf. \\  \hline & \\[-2ex]
        Climate-Stance \citep{vaid2022towards} & Stance classification & Climate-Stance \\ \\[-2ex]
        Climate-Eng \citep{vaid2022towards} &Topic classification&  Climate-Eng  \\ \\[-2ex]
        Climate-FEVER  \citep{diggelmann2020climate} &  Claim verification & Climate-FEVER\\ 
        \hline \\[-2ex]
        \multirow{3}{*}{SciDCC \citep{mishra2021neuralnere}} 
            & Topic classification by Title & SciDCC Title \\
            & Topic classification by Title \& Summary  & SciDCC Title Sum. \\
            & Topic classification by Title \& Body  & SciDCC Title Body  \\ \hline & \\[-2ex]
        \multirow{5}{*}{CLIMA-CDP \citep{spokoyny2023towards}} 
            & QA-Cities (answer relevance) & QA-Cities \\
            & QA-Corporations (answer relevance)  &  QA-Corps. \\
            & QA-States (answer relevance)  &  QA-States \\
            & QA-Topic-Cities (topic classification) & QA-Topic-Cities \\ \hline & \\[-2ex]
        \multirow{2}{*}{PIRA 2.0 MCQ \citep{pirozelli2024benchmarks}} 
            & PIRA with Context  & PIRA w/ Ctx. \\
            & PIRA without Context & PIRA w/o Ctx. \\ \hline & \\[-2ex]
        \multirow{2}{*}{Exeter Misinformation \citep{coan2021computer}}
            & Claim Detection  & Exeter Claim \\
            & Sub-claim Detection   & Exeter Sub-claim \\ \hline & \\[-2ex]
        Climate-Change NER \citep{bhattacharjee2024indus}& Entity recognition & Climate NER \\ \hline & \\[-2ex]
        \multirow{5}{*}{CheapTalk \citep{bingler2023cheaptalk}} & Climate Detection &\\ 
              & Climate Sentiment & \\
              & Climate Commitment & Climate Commit. \\
              & Climate Specificity & \\ 
              & TCFD Recommendations  & TCFD Recommend.\\\hline & \\[-2ex]
        Net-Zero Reduction \citep{schimanski2023climatebert} &Paragraph classification  & Net-Zero Reduction \\ \hline & \\[-2ex]
        Environmental Claims \citep{stammbach2022environmentalclaims} & Sentence classification & Env. Claims\\ \hline & \\[-2ex]
        \multirow{2}{*}{Guardian Climate News Corpus}  
            & Topic classification by Title & Guardian Title\\ 
            & Topic classification by Body & Guardian Body \\
        \bottomrule
    \end{tabular}
    \caption{Shortened task identifiers (if they exist) for each task presented in the Climate\-Eval benchmark. These shortened task names are used in tables presenting the results for the purpose of saving space.}
        \label{tab:climateEvalShortened}
\end{table*}

\end{document}